\documentclass{article}
\usepackage[english]{babel}
\usepackage[utf8]{inputenc}
\usepackage{amsfonts} 
\usepackage{amssymb}
\usepackage{authblk}
\usepackage[numbers]{natbib}
\usepackage{comment}
\usepackage{mathtools}
\usepackage{color}
\usepackage{booktabs}
\usepackage{algorithm}
\usepackage{algpseudocode}
\usepackage{longtable} 
\usepackage{rotating}
\usepackage{amsbsy}
\usepackage{arydshln}
\usepackage{graphicx}
\usepackage[a4paper]{geometry}
\usepackage{ccicons}

\begin{document}

\title{A Multiperiod Workforce Scheduling and Routing Problem with Dependent Tasks\thanks{\ccbyncnd 2020. This manuscript version is made available under the CC-BY-NC-ND 4.0 license http://creativecommons.org/licenses/by-nc-nd/4.0/}}
\author{Dilson Lucas Pereira\thanks{Corresponding Author: dilson.pereira@ufla.br}}
\author{Júlio César Alves \thanks{juliocesar.alves@ufla.br}}
\author{Mayron César de Oliveira Moreira \thanks{mayron.moreira@ufla.br} }
\affil{Departamento de Ciência da Computação\\Universidade Federal de Lavras}
\date{December, 2019 \\ - \\
Published version:   
https://doi.org/10.1016/j.cor.2020.104930}

\maketitle

\section*{Abstract}
In this paper, we study a new Workforce Scheduling and Routing Problem, denoted Multiperiod Workforce Scheduling and Routing Problem with Dependent Tasks. In this problem, customers request services from a company. Each service is composed of dependent tasks, which are executed by teams of varying skills along one or more days. Tasks belonging to a service may be executed by different teams, and customers may be visited more than once a day, as long as precedences are not violated. The objective is to schedule and route teams so that the makespan is minimized, i.e., all services are completed in the minimum number of days. In order to solve this problem, we propose a Mixed-Integer Programming model, a constructive algorithm and heuristic algorithms based on the Ant Colony Optimization (ACO) metaheuristic. The presence of precedence constraints makes it difficult to develop efficient local search algorithms. This motivates the choice of the ACO metaheuristic, which is effective in guiding the construction process towards good solutions. Computational results show that the model is capable of consistently solving problems with up to about 20 customers and 60 tasks. In most cases, the best performing ACO algorithm was able to match the best solution provided by the model in a fraction of its computational time.

\noindent {\bf Keywords:} Workforce Scheduling and Routing, Scheduling, Vehicle Routing

\section{Introduction}
Workforce Scheduling and Routing Problems (WSRPs) are a class of problems that involve the scheduling of service personnel for the execution of tasks at different locations \cite{castillo2012survey}. This type of problem has a broad range of applications: healthcare, equipment maintenance, building and residential construction, forestry, telecommunications, and others. Depending on the context, WSRPs may involve constraints such as time windows, special skill requirements for the execution of tasks, and dependencies among tasks. More details about these applications can be seen in \cite{begur1997integrated,castillo2016workforce,franz1989mathematical}.

This paper aims to contribute to the WSRP literature by studying a new problem, which we denote Multiperiod Workforce Scheduling and Routing Problem with Dependent Tasks (MWSRPDT), described as follows. A company executes services on customers' locations by means of mobile teams. Each customer requires the execution of a service. A service is composed of one or more tasks. There may be dependencies, or precedence relationships, among tasks, which means that some tasks may start only after the completion of others. Customers must be served over the span of one or more days. At each visit, at least one task must be executed to completion at the customer. The time taken for the completion of a task depends on the customer, the team performing the task, and the task itself. Different teams may perform tasks on a service requested by a customer and a customer may be visited more than once on a given day. Work regulations specify a maximum duration on the teams' work days. The company has to assign tasks to teams over the days, and devise routes for the teams to follow each day so that the makespan is minimized, i.e., all tasks belonging to all customers' requested services are completed in the minimum number of days. 

In order to solve the MWSRPDT, we propose a Mixed-Integer Programming (MIP) model and heuristic algorithms. The first heuristic algorithm is a constructive heuristic. Due to the precedence constraints and the possibility of customers receiving multiple visits a day, developing efficient local search algorithms for the MWSRPDT is not easy. For this reason, the Ant Colony Optimization (ACO) metaheuristic is suitable to solve the problem, as it guides constructive heuristics, not being dependent on local search algorithms. We develop and test different ACO algorithms, based on the main ACO variations and four possibilities for modeling problem components.

The remaining of the paper is organized as follows. In the next two subsections, we present the formal problem statement and review the WSRP literature. Section \ref{sec:model} introduces the mathematical model for the MWSRPDT. Section \ref{sec:heuristic} presents the heuristic approaches developed to solve the problem: the constructive heuristic is presented in Subsection \ref{subsec:constructive} and the ACO algorithms in Subsection \ref{subsec:ant}. Computational experiments are presented in Section \ref{sec:comp}: we introduce a set of problem instances in Subsection \ref{subsec:ig} and evaluate the effectiveness of the proposed algorithms in Subsection \ref{subsec:ge}. Section \ref{sec:conclusions} ends the paper with some concluding remarks.

\subsection{Formal Problem Definition}
\label{subsec:formal_definition}

A company offers a set $\mathcal{S}$ of services to its customers. Services in $\mathcal{S}$ will be represented by the letter $S$. Each service $S$ is represented by a digraph $S=(V_S, A_S)$, where $V_S$ is the set of tasks that compose the service and $A_S$ is a set of dependencies, or precedence relationships, among tasks. An arc $(a, b) \in A_S$ indicates that the execution of task $a$ can only start after the completion of task $b$. 
Letters $a$ and $b$ will be used to refer to tasks/vertices in $V_S$.
The company has $K$ teams available for the execution of the services' tasks.

The problem is modeled on a complete graph $G = (V, E)$, $|V| = n$. A particular vertex, denoted by 1, represents a depot. At the beginning of each day, teams are located at the depot, where they must return at the end of the day. Vertices $V \setminus \{1\}$ represent customers. Letters $i$ and $j$ will be used to refer to vertices in $V$.

Tasks are to be performed over a sequence of days $1, 2, \ldots$, whose length is not fixed in advance. Each customer $i$ requires the execution of a service $S^i$. All tasks composing that service must be executed someday, keeping in mind the tasks' dependencies. Different teams may perform tasks belonging to a customer's requested service, but each task must be assigned to exactly one team. A customer may be visited more than once on a given day, by the same or different teams. For simplicity, a task $a$ belonging to the service $S^i$ requested by customer $i$ will be referred to as task $a$ of customer $i$.

Each workday consists of $D \in \mathbb{R}_+$ units of time. We assume that all teams use the same means of transportation, thus it takes $d_{ij} = d_{ji} \in \mathbb{R}_+$ units of time to go from vertex $i$ to vertex $j$. On the other hand, teams are heterogeneous regarding their skills to perform tasks. In order to complete task $a$ of customer $i$, team $k$ takes $t^k_{ia} \in \mathbb{R}_+$ units of time.

The problem's objective is to schedule tasks to be executed by each team on each day and to design the routes for the visits of each team to customers, such that the number of days to complete all services is minimized.

\subsection{Literature Review}
\label{subsec:literature}

WSRPs are identified in the literature by different correlated problems. \citet{alfares2004survey} presents a survey for the Employee Tour Scheduling Problem, which aims at determining the daily and weekly plan for the workforce. The author considers the papers published from 1990 to 2004, and classify their solution techniques in 10 categories: (1) manual solution, (2) integer programming, (3) implicit modeling, (4) decomposition, (5) goal programming, (6) working set generation, (7) LP-based solution, (8) construction and improvement, (9) metaheuristics, and (10) other methods. The author points out that potential contributions could be made by addressing flexible schedules, irregular work hours, and shorter workweeks. 

Another survey on correlated problems is due to \citet{castillo2016workforce}. It is divided in two parts. The first one presents the main characteristics found in workforce scheduling and routing problems: time windows, transportation means, start and end locations, skilled workforce, tasks' service times, and precedence relationships. In the second part of the paper, the authors conduct an experimental analysis of different MIP models with 375 instances across 5 data sets. As future trends,  the study indicates the inclusion of the following features in the problems:
employees maximum workload, presence of breaks over the journey, and route balancing concerning the number of tasks.  

\citet{khalfay2017review} focus on five problems which compose the WSRP family: the Technician Task Scheduling Problem (TTSP), the Technician Routing and Scheduling Problem (TRSP), the Service Technician Routing and Scheduling problem (STRSP), the Multi-period Field Service Routing Problem (MFSRP), and the Multi-period Technician Routing and Scheduling Problem (MTRSP). According to their observation, to make an impact in the industry, future research should focus on multi-period variants, technician unavailability, strategies to build teams, and precedence constraints.

The TTSP was introduced in 2007 by the French Operational Research Society (ROADEF), as part of a challenge supported by France Telecom. In this problem, a set of tasks has to be executed by teams of technicians over a planning horizon. There are precedence relationships between tasks and in order to perform a task, a team must meet some skill requirements. Teams are to be formed by combining technicians at the beginning of each day as to meet task's requirements. The time to perform a task does not depend on the team, only on the task itself. There is also the possibility of outsourcing tasks. Tasks are partitioned into priority categories. The goal is to simultaneously determine the teams' configurations and scheduling of tasks in order to minimize the weighted makespan of the categories. This problem does not involve vehicle routing.
\citet{cordeau2010scheduling} propose a mixed-integer model, a constructive heuristic, and an adaptation of the Simulated Annealing metaheuristic. They introduce improvement methods based on Large Neighborhood Search, defined by constructive and destructive procedures. \citet{estellon2009high} follow the same research line proposed by the previous work, and develop heuristics and local search algorithms. \citet{firat2012improved} propose a MIP based heuristic approach.

\citet{solomon1987algorithms} faces a problem with routing and scheduling decisions. The author highlights specific applications, such as bank deliveries, postal deliveries, industrial refuse collection, dial-a-ride service, and school routing. In all these contexts, time-windows compose the set of constraints. That study describes several heuristics and compares their performances through extensive computational experiments. \citet{tang2007scheduling} introduce the Multiple Tour Maximum Collection Problem with Time-Dependent rewards (MTMCPTD). The problem comes from the difficulties faced by the United Technologies Corporation (UTC) to assign to teams tasks such as heating, ventilation, and air conditioning installations.  The authors maximize the total collected reward, allowing a variety of tours over the days with a limited daily journey. A mathematical model and a Tabu Search procedure are proposed for this variant.

\citet{pillac2011technician} and \citet{pillac2013parallel} consider an extension of the TRSP with time-windows and technicians with heterogeneous skills. Constructive heuristics and an adaptation of Adaptive Large Neighborhood Search are developed. In particular, \cite{pillac2013parallel} applies parallel computing and a mathematical programming based post-optimization procedure, obtaining gaps of less than 1\% in the Solomon's classical instances. \citet{chen2016technician} explicitly model workforce heterogeneity and innovate by considering learning curves based on a Markov decision process. \citet{mathlouti2018metaheuristic} describe a problem in which, in addition to having time windows, tasks require that technicians possess certain skills or spare parts for their execution. Teams are heterogeneous concerning skills and can take three breaks in a workday. The objective is to maximize the total collected gain associated with tasks, minus the total traveled distance and total overtime. The authors propose a Tabu Search scheme and test it across the benchmark proposed in \cite{mathlouthi2018mixed}. Experiments showed that this algorithm obtained the optimum for instances with less than 50 tasks, and was also able to find feasible solutions for instances up to 200 tasks. A recent paper of \citet{anoshkina2019technician} address a problem involving technician teaming and routing, with service, cost and fairness objectives. Mathematical models and bi-level decomposition schemes are proposed.

\citet{dohn2009manpower} solve the Manpower Allocation Problem with Time Windows, Job-Teaming Constraints, and a limited number of teams ($m$-MAPTWTC). In this variant, both the tasks and the technicians have specific time windows. A task might require the cooperation of teams for its execution, i.e., it might have to be simultaneously executed by different skilled teams. The goal is to maximize the number of tasks performed overall. For this purpose, the authors present a Branch-and-Price algorithm. The algorithm found optimal solutions for 11 out of 12 full-size realistic problems inspired by problems faced in airports. \citet{kovacs2012adaptive} approach the STRSP, with and without team building, with a model and a Large Neighborhood Search algorithm. The authors use real-world problem instances that involve lunch break requirements and shift length related costs.

Multiperiod planning demands attention from the literature, since it closely reflects problems faced by companies. \citet{tricoire2013exact} consider a multiperiod, multidepot uncapacitated vehicle routing problem with specific constraints, called the Multiperiod Field Service Routing Problem (MPFSRP). Several heuristics and a Branch-and-Price algorithm are tested with realistic data adapted from an industrial application. \citet{bostel2008multiperiod} deal with a problem in the water distribution sector. The problem requires the satisfaction of client demands while minimizing transportation distances, subject to time windows, work day duration, and lunch breaks. A memetic algorithm and a column generation procedure (for small instances) are proposed.

\citet{rasmussen2012home} present the Home Care Crew Scheduling Problem (HCCSP), a generalization of an uncapacitated, multi-depot Vehicle Routing Problem with Time Windows (VRPTW). The goal is to maximize the service level by minimizing uncovered visits, maximizing caretaker-visit preferences, and minimizing the total traveling costs. 
Good quality solutions are obtained by enforcing preference-based visit clustering and temporal dependencies inside a Branch-and-Price framework. \citet{zamorano2017branch} propose a Branch-and-Price algorithm for the Multiperiod Technician Routing and Scheduling Problem (MPTRSP). The MPTRSP has time window constraints, team building requirements, and precedence between tasks. Considering instances adapted from the VRPTW literature, the proposed algorithm is considered the best known in the MPTRSP literature. \citet{chen2017multi} include worker experience-based service times and stochastic customers in a multiperiod variant of the WSRP. The authors solve this problem by a Markov decision process and a dynamic programming algorithm.

\citet{zamorano2018task} investigate a task assignment problem in the context of check-in counter at airports. They consider a multiskilled workforce and outsourcing. The objective is to minimize the weighted sum of traveling time, overtime, delay, and outsourced time. The authors propose a Branch-and-Price algorithm and test its performance on real-world data from a German ground-handling agency. \citet{algethami2019adaptive} design an adaptive multiple crossovers genetic algorithm (AMCGA) with six genetic operators for the WSRP, minimizing operational and penalty costs. The rationale behind the AMCGA is to apply adaptive allocation rules regarding problem-specific and traditional crossovers, which are evaluated to measure their effectiveness. In a more flexible approach, \citet{mosquera2019flexible} study a variant of home care scheduling, taking into account client homes and caregiver depots. This problem considers caregiver preferences and qualifications, client preferences, continuity of care, fairness, flexible duration, idle time, multiple tasks, spreading, task rejection, and time windows as hard constraints. The authors consider nine lexicographic evaluation functions. Heuristic algorithms are proposed, and tests performed with Flemish home care organizations attest to the quality of the methods.

The tour scheduling problem, as defined by \citet{alfares2004survey}, despite the name, is not concerned with vehicle routing. In their terminology, a tour refers to the schedules of the employees along the week. Their problem is concerned with the assignment of workers to time slots along the week, and not the particular visitation of tasks.
The problem being studied here fits under the umbrella of WSRPs defined by \citet{castillo2016workforce} and \citet{khalfay2017review}. These works describe a list of characteristics that combined will define specific WSRPs. To the best of our knowledge, the specific combination of attributes in the problem studied here is not found in any other problem in the literature. None of the works in the literature matches the specific set of attributes present in our problem: minimization of the makespan, routing of the teams, precedence constraints, multiple periods. This specific combination of attributes is found in many real world applications: road maintenance, gardening, civil construction, telecommunications, and other industries. Among the problems in the literature, the ones that more closely match the problem studied here are the MTRSP \cite{zamorano2017branch}, the TTSP \cite{cordeau2010scheduling, estellon2009high, firat2012improved}, and the WSRP defined by the more general model in \cite{castillo2016workforce}. The first does not consider precedence constraints or the makespan in the objective function. The second does not consider routing of the teams or team dependent service times. The third is not multiperiod and does not consider the makespan in the objective function. 

Table \ref{tab:VRPFeatures} summarizes the main attributes of the WSRP variants found in the literature.

\begin{sidewaystable}[htbp]
\centering
\small
\setlength\tabcolsep{4pt} 
\begin{tabular}{@{}lcccccccccl@{}}
\toprule
 & \multicolumn{9}{c}{Attributes} & \\ 
 \cmidrule{2-10}
Reference & Multiperiod & T.W. & T.D. & Het. S. & Hom. S. & T.B. & Precedence & Outsourcing & C.A. & Contributions \\ 
\midrule
\citet{alfares2004survey} & \checkmark & \checkmark & \checkmark & \checkmark & \checkmark & \checkmark & & & & Survey (sol. methods) \\ 
\citet{algethami2019adaptive} & & \checkmark & \checkmark & \checkmark & & & & & \checkmark & Genetic Algorithm\\
\citet{anoshkina2019technician} &  &  & \checkmark & \checkmark & & \checkmark & & & & Bi-level decomposition \\
\citet{bostel2008multiperiod} & \checkmark & \checkmark & \checkmark &  & \checkmark &  &  & &  & Hybrid \\ 
\citet{castillo2016workforce} & & \checkmark & \checkmark & \checkmark & \checkmark & \checkmark & \checkmark & \checkmark & \checkmark & Survey (WSRP features and sol. methods) \\
\citet{chen2016technician} & \checkmark & \checkmark & \checkmark & \checkmark &  &  &  &  & & Hybrid \\ 
\citet{chen2017multi} & \checkmark & & & \checkmark & & & & & & Dynamic programming, experience-based\\
                      &                         & & &                         & & & & & &  service times and stochastic customers\\
\citet{cordeau2010scheduling} & \checkmark &  &  & \checkmark &  & \checkmark & \checkmark & \checkmark & \checkmark & ALNS \\ 
\citet{dohn2009manpower} &  & \checkmark & \checkmark & \checkmark &  & \checkmark &  & & \checkmark & Branch-and-Price \\
\citet{estellon2009high} & \checkmark &  &  & \checkmark &  & \checkmark & \checkmark & \checkmark & \checkmark & Heuristics \\ 
\citet{firat2012improved} & \checkmark &  &  & \checkmark &  & \checkmark & \checkmark & \checkmark & \checkmark & Hybrid \\ 
\citet{khalfay2017review} & \checkmark & \checkmark & \checkmark & \checkmark & \checkmark & \checkmark & \checkmark & \checkmark & \checkmark & Survey (WSRP features and sol. methods) \\
\citet{kovacs2012adaptive} &  & \checkmark & \checkmark & \checkmark &  & \checkmark &  & & & ALNS \\
\citet{mathlouti2018metaheuristic} &  & \checkmark & \checkmark & \checkmark & & \checkmark &  & &  & Tabu Search \\
\citet{mosquera2019flexible} & \checkmark & \checkmark & \checkmark & \checkmark & & & & & \checkmark & Hybrid metaheuristic\\
\citet{pillac2011technician} &  & \checkmark &  & \checkmark &  &  &  &  & & ALNS \\
\citet{pillac2013parallel} &  & \checkmark & & \checkmark &  &  &  &  & & Hybrid \\
\citet{rasmussen2012home} & \checkmark & \checkmark & \checkmark & \checkmark &  & \checkmark &  & & \checkmark & Branch-and-Price \\
\citet{solomon1987algorithms} &  & \checkmark &  \checkmark & & \checkmark &  &  & &  & Heuristics \\ 
\citet{tang2007scheduling} & \checkmark & & \checkmark &  & \checkmark & &  & & & Tabu Search \\ 
\citet{tricoire2013exact} & \checkmark & \checkmark & \checkmark &  & \checkmark &  & & &  & Branch-and-Price \\ 
\citet{zamorano2017branch} & \checkmark & \checkmark & \checkmark & \checkmark &  & \checkmark & & & & Branch-and-Price \\
\citet{zamorano2018task} & & \checkmark & \checkmark & \checkmark & & \checkmark & & \checkmark & & Branch-and-Price\\
\textbf{MWSRPDT} & \pmb{\checkmark} &  & \pmb{\checkmark} & \pmb{\checkmark} &  &  & \pmb{\checkmark} &  & \pmb{\checkmark} & MIP model and ACO\\
\bottomrule
\end{tabular}
	\caption{Characteristics of Workforce Scheduling and Routing Problems found in the literature.\textsuperscript{*}}
	\footnotesize{\textsuperscript{*} Legend: T.W. -- Time Windows; T.D. -- Travel Distance; Het. S. -- Heterogeneous Skills; Hom. S. -- Homogeneous Skills; T.B. -- Team Building; C.A. -- Connected Tasks (Precedence/Synchronization).}
\label{tab:VRPFeatures}
\end{sidewaystable}

\section{Mixed-Integer Programming Model}
\label{sec:model}

In this section, we introduce a Mixed-Integer Programming model for the MWSRPDT.
To simplify the exposition, we use an extended graph $G' = (V', E')$.
The vertex set $V'$ is given by $V'=\{(i,a): i \in V\setminus\{1\}, a \in S^i\} \cup \{1\}$, i.e., each customer-task pair is represented by a vertex in $G'$, plus the depot. Vertices in $V'$ will be referred to by letters $u$ and $v$. The edge set $E'$ is complete.

Let $H \in \mathbb{N}$ be an upper bound on the number of days needed to complete all requested services. This can be part of the problem's data or computed during a preprocessing phase. In our implementation, the upper bound provided by a heuristic algorithm is used to set $H$.
The following decision variables are employed in the model: 
\begin{itemize}
\item $x = \{ x_{uv}^{kh} \in \{0, 1\}: u \neq v \in V', 1 \leq k \leq K, 1 \leq h \leq H \}$. Variable $x_{uv}^{kh}$ assumes value 1 if team $k$ goes from $u$ to $v$ on day $h$, 0 otherwise. Assuming $u = (i, a)$ and $v = (j,b)$, $x_{uv}^{kh} = 1$ means that task $b$ of customer $j$ is executed right after the completion of task $a$ of customer $i$ by team $k$ on day $h$.
\item $q = \{ q_{uv}^{kh} \in \mathbb{R}_+: u \neq v \in V', 1 \leq k \leq K, 1 \leq h \leq H \}$. If team $k$ goes from $u$ to $v$ on day $h$, the value of variable $q_{uv}^{kh}$ represents the moment in which the team arrives at $v$. Assuming $v = (j,b)$, this is the moment in which team $k$ starts the execution of task $b$ of $j$.
\item $y = \{y_{ia}^{kh} \in \{0, 1\}: i \in V, a \in S^i, 1 \leq k \leq K, 1 \leq h \leq H\}$. Variable $y_{ia}^{kh}$ assumes value 1 if task $a \in S^i$ is executed on customer $i$ by team $k$ on day $h$, 0 otherwise.
\item $p \in \mathbb{Z}_+$. This variable indicates the day in which the last task is completed.
\end{itemize}

In the extended graph, travel times $d$ are also extended. The time taken by a team to go from $u \in V'$ to $v \in V'\setminus\{u\}$ is given by:
\begin{equation*}
d'_{uv} =
\begin{cases}
d_{1j} & \text{if } u = 1, v = (j, b), j \in V \setminus\{1\}, b \in S^j,\\
d_{i1} & \text{if } u = (i, a), i \in V \setminus\{1\}, a \in S^i, v = 1,\\
0 & \text{if } u = (i, a), i \in V \setminus\{1\}, a \in S^i, v = (i,b), b \in S^i\setminus\{a\},\\
d_{ij} & \text{if } u = (i, a), i \in V \setminus\{1\}, a \in S^i, v = (j,b), j \in V \setminus\{1,i\}, b \in S^j.\\
\end{cases}
\end{equation*}

The model is defined as follows. The objective function asks for the minimization of the total time to complete all requested services:
\begin{equation}
\min p.
\label{eq:obj}
\end{equation}
The constraint below grants that each customer's task is executed someday:
\begin{equation}
\sum_{k = 1}^{K}\sum_{h =1}^{H} y^{kh}_{ia} = 1, \qquad i \in V\setminus\{1\}, a \in S^i.
\end{equation}
The next constraint states that a task can only be executed if the tasks on which it depends have already been executed:
\begin{multline}
\sum_{k =1}^{K}\sum_{h = 1}^{H} (D(h-1) y^{kh}_{ia} + \sum_{u \in V'\setminus\{(i,a)\}} q^{kh}_{u(i,a)}) \geq \\
\sum_{k =1}^{K}\sum_{h = 1}^{H} ((D(h-1) + t^k_{ib})y^{kh}_{ib} + \sum_{u \in V'\setminus\{(i,b)\}} q^{kh}_{u(i,b)}), \\i \in V\setminus\{1\}, (a, b) \in A_{S^i}.
\end{multline}
Correct values for variables $q$ are granted by the following flow conservation constraints:
\begin{equation}
d'_{1v}x^{kh}_{1v} \leq q^{kh}_{1v}, \qquad v \in V'\setminus\{1\}, 1 \leq k \leq K, 1 \leq h \leq H,
\end{equation}
\begin{equation}
\sum_{u \in V'\setminus\{v\}}(q^{kh}_{uv} + d'_{vu}x^{kh}_{vu}) + t^k_{ia}y^{kh}_{ia} \leq \sum_{u \in V'\setminus\{v\}}q^{kh}_{vu}, \qquad v=(i,a) \in V'\setminus\{1\}, 1 \leq k \leq K, 1 \leq h \leq H.
\end{equation}
The constraints above are inequalities, allowing teams to wait for a previous team to finish a dependency before starting its task on that customer.
The next constraint prohibits teams from using more time than available:
\begin{equation}
q^{kh}_{uv} \leq Tx^{kh}_{uv}, \qquad u \neq v \in V', 1 \leq k \leq K, 1 \leq h \leq H.
\end{equation}
Whenever $y_{ia}^{kh} = 1$, the corresponding team should visit vertex $(i, a)$, which is granted by the following constraints:
\begin{equation}
\sum_{u \in V' \setminus\{v\}} x^{kh}_{uv} = y_{ia}^{kh}, \qquad v = (i, a) \in V'\setminus\{1\}, 1 \leq k \leq K, 1 \leq h \leq H,
\end{equation}
\begin{equation}
\sum_{u \in V' \setminus\{v\}} x^{kh}_{vu} = y_{ia}^{kh}, \qquad v = (i, a) \in V'\setminus\{1\}, 1 \leq k \leq K, 1 \leq h \leq H.
\end{equation}
No team can return to the depot and leave again on the same day:
\begin{equation}
\sum_{v \in V'\setminus\{1\}} x^{kh}_{1v} \leq 1, \qquad 1 \leq k \leq K, 1 \leq h \leq H.
\end{equation}
The day in which the last service is completed is obtained by the following constraint:
\begin{equation}
p \geq \sum_{v \in V'\setminus\{1\}} hx^{kh}_{1v}, \qquad 1 \leq k \leq K, 1 \leq h \leq H.
\label{eq:p}
\end{equation}

In Section \ref{sec:comp}, we conduct experiments to evaluate the model. It will be seen that using the model is advisable only for smaller instances. In order to solve the problem in more practical situations, heuristic algorithms might be necessary. In the next section, we develop a constructive algorithm and algorithms based on the ACO framework for the MWSRPDT.

\section{Heuristic Algorithms}
\label{sec:heuristic}

\subsection{Constructive Heuristic}
\label{subsec:constructive}

In this section, we introduce a constructive heuristic for the MWSRPDT. The development of algorithms for such a problem is tricky since when assigning tasks to teams and deciding the sequence of tasks to be executed by each team, one has to carefully determine the moments in which teams start and finish executing tasks, in order to obey the precedence constraints. The fact that many teams might visit customers on a given day makes this more complicated. In order to correctly generate feasible solutions, our algorithm, presented in Algorithm \ref{alg:in}, draws inspiration from Discrete Event Simulation \cite{leemis2006discrete}. 

The rationale behind the heuristic's functioning is as follows. Teams will follow a rule for selecting available tasks. An event queue is employed to simulate the events in the construction process as if they were happening in real-time. The next event drawn from the queue is always the one that happens the earliest. An event indicates that a team has just finished a task, and is ready to select another task according to the selection rule. Thus, when a team completes a task, we know precisely the tasks that are already available,  the ones that are not available yet but the moment when they will become available is known, and the ones which we can not be sure. The team then applies the selection rule to select a task from one of the first two groups. The moment in which the team will start executing that task can be determined by examining the time taken to travel from its current position to the customer to which that task belongs and the moment in which that task became or will become available. The moment in which the team will finish the task is obtained by adding the time taken to execute the task to the starting moment. This information is used to update the dependent, or successor, tasks. A new event is then added to the event queue, and scheduled to happen when that team completes the task. Following this process, we can be sure that no precedence relationship will be broken.

In step 25 of Algorithm \ref{alg:in}, teams choose from available tasks. The implementation of this step depends on the context in which the algorithm is being applied: individually, during the exploration of the search tree when the model is being solved, or as the basis for the ACO algorithms discussed in the next section.

When Algorithm \ref{alg:in} is being used by itself, teams will select the task $(j,b)$ with the smallest $\mathtt{end}[j,b]$.  
In an attempt of finding good upper bounds when exploring the search tree, the algorithm is also used in conjunction with model \eqref{eq:obj}-\eqref{eq:p}. In this case, the algorithm is called several times during the search tree exploration. Let $(\overline{x}, \overline{y}, \overline{q}, \overline{p})$ be the possibly fractional solution of current search tree node when Algorithm \ref{alg:in} is invoked. Then, in step 25, teams will select the available task $(j,b)$ for which the sum $\overline{x}^{kh}_{(i,a)(j,b)} + \overline{y}^{kh}_{jb}$ is maximum. 

The third context will be explained in the next section, in which Algorithm \ref{alg:in} is used as the basis for ACO algorithms.

\begin{algorithm}[htpb]
\caption{MWSRPDT constructive heuristic.}
\begin{algorithmic}[1]
\State $\mathtt{currentDay} \leftarrow 0$
\State $\mathtt{solutionComponents} \leftarrow \emptyset$
\State Let {\tt toExecute} be the set of all pairs $(i, a)$ where $i$ is a customer and $a$ is a task from the service requested by that customer.
\For{all $(i,a)$ in $\mathtt{toExecute}$}
    \State Let $\mathtt{open}[i,a] \leftarrow (0,0)$ if $a$ has no dependencies, and $(\infty, \infty)$ otherwise. This pair represents the moment in which task $a$ of customer $i$ becomes available, i.e., the earliest moment in which all of its dependencies where in a completed state. The first component of the pair is an integer number representing the day and the second component is a real number in $[0, D]$, representing the moment within the day in which such event happens.
    \State Likewise, let $\mathtt{completed}[i,a] \leftarrow (\infty, \infty)$ indicate the moment, i.e., the day and moment within the day, in which the task is completed.
\EndFor
\While{$\mathtt{toExecute} \neq \emptyset$}
    \State Increment {\tt currentDay}
    \State Let {\tt available} be the set of all pairs $(i, a)$ in $\mathtt{toExecute}$ such that $\mathtt{open}[i,a] \neq (\infty, \infty)$.
    \State Let {\tt events} be an event queue with tuples $(k, i, a, q)$, representing tasks being performed, where $k$ is a team, $i$ is a vertex, $a$ is a task from that vertex's service ({\tt null} if $i = 1$), and $q$ is the moment in which that task will be completed (0 if $i = 1$). The queue is organized such that the next tuple to be removed is always the one with least $q$.
    \State Initialize {\tt events} with $(k, 1, \mathtt{null}, 0)$ for all teams $k$.
    \While{$\mathtt{events} \neq \emptyset$}
        \State Remove the next event $(k, i, a, q)$ from {\tt events}.
        \For{each $(j,b)$ in $\mathtt{available}$}
            \State Let $s$ be the second component of $\mathtt{open}[j,b]$ if the first component is equal to $\mathtt{currentDay}$ and 0 if it is less than $\mathtt{currentDay}$.
            \State Let $\mathtt{start}[j,b]$ be earliest possible moment in which team $k$ could start performing that task, i.e., $\mathtt{start}[j,b] \leftarrow \max\{ q + d'_{(i,a)(j,b)}, s\}$.
            \State Let $\mathtt{end}[j,b]$ be the moment in which team $k$ would complete that task if it were to start it in $\mathtt{start}[j,b]$, i.e., $\mathtt{end}[j,b] \leftarrow \mathtt{start}[j,b] + t_{jb}^k$.
        \EndFor
        \State Let {\tt availableForTeam} be the set of all pairs $(j, b)$ in $\mathtt{available}$ that can still be executed by $k$ in the current day, i.e., $\mathtt{end}[j,b] + d'_{(j,b)1} \leq D$.
        \If{$\mathtt{availableForTeam} = \emptyset$}
            \State Team $k$ returns to the depot.
            \State Add the corresponding component to {\tt solutionComponents}
        \Else
            \State Pick a task $(j,b)$ from $\mathtt{availableForTeam}$ to be executed by team $k$, starting in $\mathtt{start}[j,b]$.
            \State Add the corresponding component to {\tt solutionComponents}
            \State Remove $(j,b)$ from $\mathtt{toExecute}$ and from $\mathtt{available}$. 
            \State Set $\mathtt{completed}[j,b] \leftarrow (\mathtt{currentDay}, \mathtt{end}[j,b])$
            \For{each task $c$ that depends on $b$}
                \State Set $\mathtt{open}[j, c]$ to the maximum value of $\mathtt{completed}[j, c']$ for all $c'$ on which $c$ depends.
                \State If $\mathtt{open}[j, c] \neq (\infty, \infty)$, add $(j, c)$ to $\mathtt{available}$.
            \EndFor
            \State Add the event $(k, j, b, \mathtt{end}[j,b])$ to $\mathtt{events}$.
        \EndIf
    \EndWhile
\EndWhile
\end{algorithmic}
\label{alg:in}
\end{algorithm}

\subsection{Ant Colony Optimization Algorithms}
\label{subsec:ant}

A general approach for solving hard combinatorial optimization problems is to apply local search to initial solutions generated by constructive heuristics. Often, this process will be guided by some metaheuristic framework. Developing efficient local search procedures for the MWSRPDT turns out to be challenging. Any changes made to a solution, e.g., relocating a task in a route or assigning a task to a different team, are very likely to affect the task's starting and completion time, which might break precedence constraints. To go around this problem, we resort to Ant Colony Optimization \cite{dorigo2003ant}. Differently, from many metaheuristics, which start from a complete solution and guide the process of modifying that solution by local search or perturbation, ACO guides the construction process and is not dependent on local search. 

Constructive heuristics, such as Algorithm \ref{alg:in}, build feasible solutions one step at a time, by starting from an empty solution and iteratively selecting components to be added to the current partial solution. An ACO algorithm is built upon a constructive heuristic, guiding the construction process while simultaneously learning from the solutions it generates. The ACO mechanism draws inspiration from the natural behavior of ants to guide the selection process towards improved solutions \cite{dorigo2003ant}.

At each iteration of an ACO algorithm, agents known as ants independently apply the constructive heuristic to obtain a solution, during the so-called construction phase. Whenever a new component has to be added to the current partial solution, e.g., step 25 of Algorithm \ref{alg:in}, a non-deterministic mechanism, which favors components that are likely to appear in good solutions, is employed. After solutions are generated, information obtained from the pool of solutions generated so far is used to calibrate the selection mechanism for the next iteration of the ACO algorithm. This is known as the offline pheromone update phase.

Let $\mathcal{C}$ be the set of components that make up problem solutions. Let $C \subseteq \mathcal{C}$ be the set of components available to be selected at the current iteration of the constructive algorithm of an ant. In Algorithm \ref{alg:in}, the set of components available to be added to the current partial solution is implied by the set $\mathtt{availableForTeam}$. A component could be an arc $((i,a), (j,b))$, where $(i,a)$ is the vertex and task in which team $k$ currently stands on, and $(j,b)$ is a vertex-task pair in $\mathtt{availableForTeam}$.
The basic ACO selection mechanism dictates that components $c \in C$ should be chosen randomly according to the probability
\begin{equation}
p_c = \frac{\tau_c^{\alpha}\eta_c^{\beta}}{\sum_{c' \in C}\tau_{c'}^{\alpha}\eta_{c'}^{\beta}},
\label{eq:antprob}
\end{equation}
where $\tau_c$, known as pheromone, is a measure of attractiveness of $c$ as learned by the ACO algorithm up to that point, and $\eta_c$ is also a measure of attractiveness of $c$, but from an heuristic point of view. In our implementation, $\eta_c$ is the ending time of a task in $\mathtt{availableForTeam}$. Algorithm parameters $\alpha \in \mathbb{R}_+$ and $\beta \in \mathbb{R}_+$ are used to favor one over the other.

Initially, the pheromone of each component is set to a parameter $\tau_0 \in \mathbb{R}_+$. After the construction phase, pheromones are updated. Let $\cal Z$ denote the set of solutions that will be used for updating the pheromones. A common approach is to let $\mathcal{Z}$ be the solutions generated during the construction phase. Let the notation $c \in Z$ indicate that component $c$ is used in $Z \in \mathcal{Z}$. Given $Z \in \mathcal{Z}$, let $f(Z)$ denote its objective value, e.g., the number of days in the MWSRPDT solution represented by $Z$. A common strategy for pheromone update is to set:
\begin{equation}
\tau_c \leftarrow (1-\rho) \cdot \tau_c + \sum_{Z \in \mathcal{Z} : c \in Z}\frac{Q}{f(Z)},
\label{eq:update}
\end{equation}
where $\rho \in [0,1]$, known as evaporation rate, and $Q \in \mathbb{R}_+$, are algorithm parameters. The evaporation rate controls how fast the ACO algorithm forgets information learned in previous iterations.

In our implementation, instead of $f(Z)$ in \eqref{eq:update}, we use $f'(Z)=f(Z)-1+m/D$, where $m \in [0, D]$ is the moment in the last day of the solution represented by $Z$ in which the last task was completed. Better results were obtained by using $f'$ instead of $f$ since solutions with less total (fractional) time are favored.

There are three main ACO variations. The first is the one described above, Ant System. The second is Max-Min Ant System (MMAS). 

In MMAS, after the construction phase, pheromones are updated only by either the global best solution found thus far, or by the iteration best. MMAS also imposes upper and lower limits, $\tau_{\max}$ and $\tau_{\min}$, respectively, on the amount of pheromone $\tau_c$ on a component $c$. 

The third variation is the Ant Colony System (ACS). As in MMAS, in the offline pheromone update phase pheromones are updated only by either global or iteration best ants. Besides offline pheromone updates, ACS also introduces what is called local pheromone update, a scheme in which the pheromone of a component is updated as soon as it is selected by an ant in the construction phase:
\begin{equation}
\tau_c \leftarrow (1-\phi) \cdot \tau_c + \phi \cdot \tau_0,
\label{eq:onlineu}
\end{equation}
where $\phi \in [0,1]$ is an algorithm parameter.

Another distinctive feature of ACS is the Pseudo-random Proportional Rule for selecting components during the construction phase. In this scheme, the rule to be applied for selecting the next component to be added to the current partial solution depends on a random value $q$ uniformly distributed in $[0,1]$. Given an algorithm parameter $q_0 \in [0,1]$, if $q < q_0$, then the next component to be added to the current partial solution is the one that maximizes $\tau_c\eta_c^{\beta}$. Otherwise, the next component is selected according to the probability in \eqref{eq:antprob}.  
It is not the focus of this paper to present an in-depth discussion of the ACO metaheuristic, and we refer the interested reader to \cite{dorigo2003ant}.

We implemented the three main ACO variations, AS, MMAS, and ACS, for which pseucodes are respectively presented in Algorithms \ref{alg:as}, \ref{alg:mmas}, and \ref{alg:acs}.

\begin{algorithm}[htpb]
\caption{MWSRPDT Ant System heuristic.}
\begin{algorithmic}[1]
\State For every component $c \in \mathcal{C}$, set $\tau_c \leftarrow \tau_0$.
\For{{\tt iter} in $1 \ldots \mathtt{maxIter}$}
    \State Let $\mathcal{Z}$ be the (multi)set of solutions obtained by executing {\tt numAnts} independent instances of Algorithm \ref{alg:in}. In each execution, step 25 is replaced by:
    \State \hspace{0.3cm} {\it Step 25}: ``Randomly pick a task $(j,b)$ from $\mathtt{availableForTeam}$ to be executed by team $k$, starting in $\mathtt{start}[j,b]$, according to probability \eqref{eq:antprob} of the associated component.'' 
    \State For each component $c \in C$, use the ants in $\mathcal{Z}$ to update $\tau_c$ according to \eqref{eq:update}.
\EndFor
\State \textbf{return} the best solution found.
\end{algorithmic}
\label{alg:as}
\end{algorithm}

\begin{algorithm}[htpb]
\caption{MWSRPDT Max-Min Ant System heuristic.}
\begin{algorithmic}[1]
\State For every component $c \in \mathcal{C}$, set $\tau_c \leftarrow \tau_0$.
\For{{\tt iter} in $1 \ldots \mathtt{maxIter}$}
    \State Let $\mathcal{Z}$ be a (singleton) set containing the best solution obtained by executing {\tt numAnts} independent instances of Algorithm \ref{alg:in}. In each execution, step 25 is replaced by: 
    \State \hspace{0.3cm} {\it Step 25}: ``Randomly pick a task $(j,b)$ from $\mathtt{availableForTeam}$ to be executed by team $k$, starting in $\mathtt{start}[j,b]$, according to probability \eqref{eq:antprob} of the associated component.'' 
    \State For each component $c \in C$, use the best ant obtained in this iteration to update $\tau_c$ according to \eqref{eq:update}. If $\tau_c > \tau_{\max}$ let $\tau_c \leftarrow \tau_{\max}$. If $\tau_c < \tau_{\min}$ let $\tau_c \leftarrow \tau_{\min}$.
\EndFor
\State \textbf{return} the best solution found.
\end{algorithmic}
\label{alg:mmas}
\end{algorithm}

\begin{algorithm}[htpb]
\caption{MWSRPDT Ant Colony System heuristic.}
\begin{algorithmic}[1]
\State For every component $c \in \mathcal{C}$, set $\tau_c \leftarrow \tau_0$.
\For{{\tt iter} in $1 \ldots \mathtt{maxIter}$}
    \State Let $\mathcal{Z}$ be a (singleton) set containing the best solution obtained by executing {\tt numAnts} independent instances of Algorithm \ref{alg:in}. In each execution, step 25 is replaced by the following two steps
    \State \hspace{0.3cm} {\it Step 25.1}: ``Draw a random number $q$ using a uniform distribution in $[0,1]$. If $q \geq q_0$ Randomly pick a task $(j,b)$ from $\mathtt{availableForTeam}$ to be executed by team $k$, starting in $\mathtt{start}[j,b]$, according to probability \eqref{eq:antprob} of the associated component. Otherwise, select the task whose associated component $c$ maximizes $\tau_c\eta_c^{\beta}$.'' 
    \State \hspace{0.3cm} {\it Step 25.2}: Right after executing step 25.1, let $c$ be the selected component and update $\tau_c$ according to \eqref{eq:onlineu}.
    \State For each component $c \in C$, use the best ant obtained in this iteration to update $\tau_c$ according to \eqref{eq:update}.
\EndFor
\State \textbf{return} the best solution found.
\end{algorithmic}
\label{alg:acs}
\end{algorithm}

Our algorithms were implemented with the help of the Formigueiro ACO framework, developed by ourselves and available as open-source code in \cite{Formigueiro}. In Formigueiro, the user selects the desired ACO variation to be used and implements a constructive heuristic. Whenever a component needs to be selected by that heuristic, the user calls a particular framework function to choose for him. Formigueiro will decide according to the scheme of the selected variation. All the steps of the ACO scheme, such as pheromone update, are handled by the framework.

The modeling of components is an important factor in ACO as it determines what is learned by the algorithm. In our implementation, we could model components as tuples $((i,a), (j,b))$, where $(i,a)$ is the current vertex-task pair of team $k$, and $(j,b)$ is a vertex-task pair in $\mathtt{availableForTeam}$, representing the arc to be followed by team $k$ in the extended graph. If that were the case, the ACO algorithm would then learn which such arcs should be included in the solution, irrespective of the day or team as that information is not encoded in the component. This component modelling approach will be denoted $ct_1$.
An alternative is to encode team information in the components, modeling components as tuples $(k, (i,a), (j,b))$. In doing so, there is a now distinction between $(k, (i,a), (j,b))$ and $(k', (i,a),(j,b))$. The former represents team $k$ following arc $((i,a),(j,b))$, while the latter represents team $k'$ following the same arc. This modelling approach will be denoted $ct_2$.
Another alternative is to also include the day $h$ in which such arcs should be traversed, and model components as tuples $(h, k, (i,a),(j,b))$. This modeling approach will be referred to as $ct_3$. 
A simpler approach, denoted $ct_4$, is to model components as tuples $(k, (i,a))$, so that the algorithm learns what tasks should be performed by each team, irrespective of the day or preceding activity.
For each of the three ACO variations we tried each of these four modelling options. Computational results involving the proposed ACO algorithms are presented in the next section.

\section{Computational Experiments}
\label{sec:comp}

In this section, we present computational experiments. We first describe the generation of problem instances. After that, we discuss the calibration of the ACO parameters and the choice of the ACO variation and component type most suitable for the MWSRPDT. Finally, we present more general experiments to evaluate the constructive heuristic, the winning ACO implementation, and the proposed model. 

\subsection{Instance Generation}
\label{subsec:ig}
To evaluate model \eqref{eq:obj}-\eqref{eq:p} and the algorithms proposed here, we generate three types of artificial instances, type A, B, and C. We first describe the characteristics that are shared by all three instance types.

Assume we are given the number $n$ of vertices, the number $\mathcal{|S|}$ of available services such that $\mathcal{S} = \{S_1, \ldots, S_{|\mathcal{S}|}\}$, the number $|V_S|$ of tasks in each service $S \in \mathcal{S}$, the number $K$ of teams, and the work day's length $D$. 

In order to generate $G$, the depot and customers are initially placed at random at the vertices of a $100 \times 100$ grid. The intention is to simulate an urban area.  
Given two vertices $i$ and $j$, respectively located in the grid with coordinates $(\mathrm{x}_i, \mathrm{y}_i)$ and $(\mathrm{x}_j, \mathrm{y}_j)$, the number of edges in the grid's shortest path between them is $|\mathrm{x}_i-\mathrm{x}_j|+|\mathrm{y}_i-\mathrm{y}_j|$.
Graph $G$ is complete, it is assumed that the teams travel at a speed of $40km/h$, and the distance between consecutive rows and column of the grid is $0.1km$. Thus, the travel time $d_{ij}$ between vertices $i$ and $j$ is $\frac{0.1(|\mathrm{x}_i-\mathrm{x}_j|+|\mathrm{y}_i-\mathrm{y}_j|)}{40}$ hours.

To generate the precedence graph of each service, we initially spread its tasks at random over three groups. Tasks from the first group have no dependencies. Every task from the second group depends on every task from the first group, and every task from the third group depends on every task from the second group. 

In type A instances, in order to determine the tasks' execution times, a reference number $r_a$ is sampled from $\{0.5, 1, 1.5, 2\}$ for each task $a$. Then, for each team $k$, a skill level $s^k_a$ for the team in that task is sampled from $\{0.5, 1, 2\}$. The time taken by team $k$ to execute task $a$ of customer $i$ is then given by $t^k_{ia} = r_a/s^k_a$.  For these instances, we let  $|\mathcal{S}| = 3$ and $K=3$. The number of tasks in services $S_1$, $S_2$ and $S_3$ is respectively set to 1, 3, and 5. Each customer requests a service selected at random.

Type B instances are similar to type A instances, except that some teams might not have the skills necessary to execute some tasks. Each task $a$ is granted to have at least one team capable of executing it, denoted $k(a)$, selected at random among the $K$ teams. The skill level for a team to execute task $a$ is then sampled from $\{0, 0.5, 1, 2\}$, if this team is not $k(a)$. For $k(a)$, the skill level is sampled from $\{0.5, 1, 2\}$. Then, the time taken by team $k$ to execute task $a$ of customer $i$ is given by $t^k_{ia} = r_a/s^k_a$, if $s^k_a \neq 0$, and $t^k_{ia} = \infty$, if $s^k_a = 0$, meaning that this team cannot execute this task. 

In type C instances, only team $k(a)$ is capable of executing task $a$. Thus, $t^k_{ia} = \infty$ if $k \neq k(a)$. For $k(a)$, $t^k_{ia} = r_a/s^k_a$, where $s^k_a$ is randomly sampled from $\{0.5, 1, 2\}$. For these instances, only one service comprising 3 tasks is available. Each one of the $K=3$ teams is responsible for a unique task.

We consider $n \in \{10, 15, 20, 25, 30, 35, 40, 50, 60, 70, 80, 90, 100\}$, and let $D=8$ hours for all instances.
The instance generation process is detailed in Algorithm \ref{alg:instance}.

\begin{algorithm}[htpb]
\caption{Instance Generation.}
\begin{algorithmic}[1]
\Require Number of vertices $n$; $\mathtt{instanceType} \in \{A, B, C\}$.
\State Set the number of teams $K$ to $3$. Set the work day's length $D$ to $8$ hours.
\State Let $G$ be a complete graph with $n$ vertices.
\State For each $i \in V$, generate random coordinates $x_i \in \{0, \ldots, 100\}$ and $y_i \in \{0, \ldots, 100\}$ with uniform probability. Then, set the travel time $d_{ij}$ to $\frac{0.1(|\mathrm{x}_i-\mathrm{x}_j|+|\mathrm{y}_i-\mathrm{y}_j|)}{40}$ for each pair of vertices $i$ and $j$.
\If{$\mathtt{instanceType} \in \{A, B\}$}
    \State There are three available services, $\mathcal{S}=\{S_1, S_2, S_3\}$, with the number of tasks in $V_{S_1}$, $V_{S_2}$, and $V_{S_3}$ respectively set to 1, 3, and 5.
\ElsIf{$\mathtt{instanceType} = C$}
    \State There is one available service, $\mathcal{S}=\{S_1\}$, comprising $|V_{S_1}|=3$ tasks. 
\EndIf
\For{ each service $S \in \mathcal{S}$}
    \State For each task $a \in V_S$, let its reference execution time $r_a$ be a random number sampled from $\{0.5, 1, 1.5, 2\}$ with uniform probability. If $\mathtt{instanceType} \in \{B, C\}$, let $k(a)$ be a random team from $\{1, \ldots, K\}$, selected with uniform probability, that is granted to be able to perform the task.
    \State Partition its set of tasks $V_S$ into three sets $V_S^1$, $V_S^2$, and $V_S^3$.
    \State Let the dependencies $A_S \leftarrow \emptyset$.
    \State For $l \in \{2, 3\}$, for each task $a$ in $V_S^l$ and $b$ in $V_S^{l-1}$, add $(a, b)$ to $A_S$.
\EndFor
\State For each $i \in V\setminus \{1\}$, let its requested service $S^i$ be selected at random from $\mathcal{S}$ with uniform probability.
\For{For each team $k \in \{1, \ldots, K\}$, and task $a \in V_S$, $S \in \mathcal{S}$}
    \If{$\mathtt{instanceType} = A$}
        \State Let the skill $s_a^k$ of $k$ in $a$ be sampled from $\{0.5, 1, 2\}$ with uniform probability. 
    \ElsIf{$\mathtt{instanceType} = B$}
        \State Let the skill $s_a^k$ of $k$ in $a$ be sampled from $\{0, 0.5, 1, 2\}$ if $k \neq k(a)$, and from $\{0.5, 1, 2\}$ otherwise.
    \Else
        \State Let the skill $s_a^k$ of $k$ in $a$ be 0 if $k \neq k(a)$, otherwise sample it from $\{0.5, 1, 2\}$ with uniform probability.
    \EndIf
    \State For each $i \in V\setminus\{1\}$ such that $S^i = S$, set the time $t^k_{ia}$ taken by team $k$ to perform task $a$ at $i$ to $r_a/s^k_a$ if $s^k_a \neq 0$, and $t^k_{ia} = \infty$ otherwise.
\EndFor
\end{algorithmic}
\label{alg:instance}
\end{algorithm}

\subsection{ACO Parametrization}
\label{sec:par}
Our first set of experiments was performed in order to determine the best parametrization, including the best choice for modeling components, for each of the ACO algorithms: AS, MMAS, and ACS.
For these experiments, we employed 10 instances of each type (A, B, and C), a total of 30 instances, all of them with $n=20$ vertices.

The algorithms were all coded in {\tt python3}, and the parameters tuned with the help of the Hyperopt python library \cite{Hyperopt}. Hyperopt is a popular tool for hyper-parameter definition in Machine Learning algorithms. 
Each ACO algorithm was run for 100 iterations with 100 ants for each particular parameter setting determined by Hyperopt, on each of the ten instances. 
Instead of looking for the parameter setting that minimized the average objective value, we instructed Hyperopt to find the parameter tuning that minimized the average ratio between the ACO solution and the constructive heuristic solution, i.e., the parameter tuning that offered the best average improvement over the constructive heuristic solution. This was done because instances of type C have significantly larger solutions, and would dominate the average objective value. Considering the average ratio to the constructive heuristic solution works as form of normalization of the objective values.
No more than than 100 different parameter settings were allowed for each algorithm (AS, MMAS, ACS).

Table \ref{tab:hyper} presents the results. Each row, except for the last, corresponds to a parameter to be tuned. Each parameter is referred to by its most common literature notation. The last row presents, for the best tuning of each algorithm, the ratio between the ACO and constructive heuristic solution averaged over the 30 instances. The column labeled ``domain'' presents the domain for the values of each parameter. Columns AS, MMAS and ACS present the best tuning found by Hyperopt, a ``--'' indicates that the parameter does not apply for that variation. The best performing algorithm was MMAS with component modelling approach $ct_3$, which on average improved the constructive heuristic solution by 17\%. This winning MMAS implementation will be used for the remaining experiments in the paper.

\begin{table}[htbp]
\centering
\begin{tabular}{@{}lcccc@{}} 
\toprule
parameter   & domain  & AS   & MMAS & ACS\\
\midrule
$\alpha$    & [0, 10] & 5.97 & 6.47 & 9.29\\
$\beta$     & [0, 10] & 1.39 & 5.78  & 0.53\\
$\rho$      & [0, 1]  & 0.48 & 0.02 & 0.82\\
$Q$         & [1, 10] & 4.08 & 9.96 & 8.91\\
$\tau_0$    & [1, 10] & 9.99 & 8.88 & 7.28\\
$\tau_{\min}$ & [0, 1]  & --   & 0.02 & --\\
$\tau_{\max}$ & [1, 10] & --   & 5.69 & --\\
$\phi$      & [0, 1]  & --   & --   & 0.12\\
$q_0$       & [0, 1]  & --   & --   & 0.91\\
    component   & $\{ct_1,ct_2,ct_3,ct_4\}$ & $ct_2$ & $ct_3$ & $ct_2$\\
obj. val.   & --      & 0.84  & 0.83 & 0.85\\
\bottomrule
\end{tabular}
\caption{Parameter setting for the ACO algorithms.}
\label{tab:hyper}
\end{table}

\subsection{General Experiments}
\label{subsec:ge}
In this section, we present the results of experiments conducted with the proposed instances of type A, B, and C, to evaluate the model and the algorithms proposed here. For each instance type and $n \in \{10, 15, 20, 25, 30, 35, 40, 50, 60, 70, 80, 90, 100 \}$, 10 instances were generated. Note that the instances with $n=20$ used in this section are not the same as those used in the paramerization of the ACO algorithms (Section \ref{sec:par}).

The algorithms were all coded in {\tt python3}. The source code and instances used in this paper can be found at {\tt github.com/dilsonpereira/MWSRPDT}. To solve the model, we employed the CPLEX optimization package, version 12.7, obtained through the IBM Academic Initiative. CPLEX's tuning parameters are all left at their default values. All the experiments were conducted on a machine running under Ubuntu 14.04 operating system, with an 8-core 3.5Ghz Intel Xeon CPU and 8GB of RAM.

Two set of experiments were conducted. In the first set, the model and heuristic algorithms are applied to small to medium-sized instances, with $n \leq 35$. In the second set, the heuristic algorithms are applied to medium to large-sized instances, with $ n \geq 40$.

Results for type A, B, and C instances in the first set of experiments are respectively presented in Tables \ref{tab:A}, \ref{tab:B}, and \ref{tab:C}. In each table, under ``Instance'' we present the number of vertices $n$,  an identifier ``id'', and the number of tasks in the instance. Under ``Constructive'' we present results for the constructive algorithm. The number of days in the solution is given under ``ub'', and the computational time in seconds to obtain the solution under ``t''. Under ``ACO'' we present results for the ACO algorithm. The number of days in the best solution found is given under ``ub'' and the total computational time in seconds to run the ACO algorithm under ``t''. The best solution found by the ACO algorithm was used to initialize the model. Results for the model are presented under ``Model''. We present the formulation's linear relaxation value, under ``LR''; the root lower bound given by CPLEX after possibly adding its cuts, under ``root''; the final upper and lower bounds obtained by executing the model with a time limit of 3600 seconds, respectively under ``ub'' and ``lb''; the total number of nodes explored, under ``nodes''; and the total computational time in seconds, under ``t''. The symbol ``*'' indicates that the model failed to close the problem within the time limit. Some of these instances resulted in problems that were too large for the solver to solve even the root node relaxation. These are indicated with the symbol ``-'' under the model entries. As only one type C instance with 15 vertices was solved to optimality, we decided to not proceed further with the experiments on type C instances. Results for the heuristic algorithms on these instances will be presented with those for the second set of experiments. For each size of $n$, we also present average results over the ten instances with that size. Instances for which the solver was unable to solve the root node relaxations are not used in the computation of these averages.

Considering type A instances, we see that the model manages to consistently solve problems with 20 vertices or less. Out of 30 such problems, 24 are solved. When $n\geq 25$, most problems cannot be solved; for these cases, it is not advisable to use the model. Another interesting fact is that most problems are solved when CPLEX applied some form of post-processing to the root nodes, as only one node was effectively explored.  We also see that the constructive algorithm is fast, but in many cases gives solutions that require one day more than the solution provided by the ACO algorithm.  In only 4 cases out of 60, the ACO algorithm failed to match the best solution given by the model. The ACO algorithm is, however, significantly faster than the model, requiring no more than a few minutes to find good solutions.

The same conclusions drawn for type A instances seem to apply for type B instances. The model consistently solved problems with 20 vertices or less. Out of 30 instances with $n \leq 20$, 20 were solved by the model. These instances seem slightly harder than type A instances.  In type B instances, the best solution found by the ACO algorithm matches the best solution found by the model in 56 out of 60 cases.  On average, the ACO algorithm finds a solution that is about one day less than the solution of the constructive algorithm. For the larger instances, there is a difference of almost two days, on average, which would be very significant in a real-world context. On instance 8 with $n=35$, the solution found by the ACO algorithm takes five days less than the solution found by the constructive algorithm.

Type C instances seem significantly harder. The model fails to solve most of even the smallest problems to optimality. Besides, the ACO algorithm does not perform as well as in the other types; it fails to match the best solution found by the model in 5 of the 20 cases. The improvement by the ACO algorithm over the constructive algorithm is also not as significant as in type A and B instances.

\begin{table}[htbp]
\centering
\scriptsize
\begin{tabular}{@{}lrrrrrrrrrrrrrrr@{}} 
\toprule
 \multicolumn{3}{c}{Instance} & & \multicolumn{2}{c}{Constructive} & & \multicolumn{2}{c}{ACO} & & \multicolumn{6}{c}{Model}\\
\cmidrule{1-3} \cmidrule{5-6} \cmidrule{8-9} \cmidrule{11-16}
    $n$ & id & tasks & & ub & t & & ub & t & & ub & lb & LR & root & nodes & t\\
\midrule 
10 & 0 & 23 &  & 2 & 0 &  & 2 & 11.6 &  & 2 & 2 & 0.65 & 1 & 1 & 8.8\\
 & 1 & 35 &  & 3 & 0 &  & 2 & 13.9 &  & 2 & 2 & 0.79 & 0.79 & 1 & 8.7\\
 & 2 & 31 &  & 2 & 0 &  & 2 & 13.1 &  & 2 & 2 & 0.82 & 0.82 & 1 & 7\\
 & 3 & 35 &  & 3 & 0 &  & 2 & 16.5 &  & 2 & 2 & 0.97 & 0.97 & 1 & 30.6\\
 & 4 & 23 &  & 2 & 0 &  & 2 & 8.8 &  & 2 & 1 & 0.6 & 1 & 24862 & *\\
 & 5 & 23 &  & 2 & 0 &  & 2 & 9 &  & 2 & 2 & 0.64 & 1 & 424 & 192.6\\
 & 6 & 29 &  & 2 & 0 &  & 2 & 13.4 &  & 2 & 2 & 0.83 & 0.83 & 1 & 20.7\\
 & 7 & 29 &  & 2 & 0 &  & 2 & 12.3 &  & 2 & 1 & 0.63 & 1 & 5301 & *\\
 & 8 & 31 &  & 2 & 0 &  & 2 & 15.5 &  & 2 & 2 & 0.78 & 0.78 & 1 & 16.6\\
 & 9 & 19 &  & 1 & 0 &  & 1 & 7.2 &  & 1 & 1 & 0.45 & 0.45 & 1 & 0.3\\
 \cmidrule{2-16}
 & average & 27.8 &  & 2.1 & 0 &  & 1.9 & 12.1 &  & 1.9 & 1.7 & 0.72 & 0.86 & 3059.4 & 748.5\\
\midrule
15 & 0 & 48 &  & 2 & 0 &  & 2 & 33.6 &  & 2 & 2 & 0.83 & 0.83 & 1 & 22.8\\
 & 1 & 48 &  & 3 & 0 &  & 2 & 24.1 &  & 2 & 2 & 1.16 & 1.16 & 1 & 18.5\\
 & 2 & 48 &  & 4 & 0 &  & 3 & 29.7 &  & 3 & 3 & 1.68 & 1.68 & 1 & 261.2\\
 & 3 & 52 &  & 3 & 0 &  & 3 & 29.6 &  & 2 & 2 & 0.92 & 1.72 & 1 & 401.2\\
 & 4 & 40 &  & 2 & 0 &  & 2 & 25 &  & 2 & 2 & 0.81 & 0.81 & 1 & 42.3\\
 & 5 & 42 &  & 3 & 0 &  & 2 & 21.8 &  & 2 & 2 & 0.97 & 0.97 & 1 & 37.7\\
 & 6 & 38 &  & 3 & 0 &  & 2 & 21.7 &  & 2 & 2 & 1.01 & 1.01 & 1 & 7.6\\
 & 7 & 46 &  & 3 & 0 &  & 3 & 26 &  & 3 & 2 & 1.09 & 2 & 453 & *\\
 & 8 & 46 &  & 3 & 0 &  & 3 & 34 &  & 3 & 3 & 1.43 & 1.43 & 1 & 246.7\\
 & 9 & 38 &  & 2 & 0 &  & 2 & 21.1 &  & 2 & 2 & 0.73 & 0.73 & 1 & 41.7\\
 \cmidrule{2-16}
 & average & 44.6 &  & 2.8 & 0 &  & 2.4 & 26.7 &  & 2.3 & 2.2 & 1.06 & 1.23 & 46.2 & 468\\
\midrule
20 & 0 & 63 &  & 5 & 0 &  & 5 & 55.6 &  & 5 & 4 & 2.5 & 4 & 30 & *\\
 & 1 & 63 &  & 5 & 0 &  & 3 & 39.3 &  & 3 & 3 & 1.88 & 1.88 & 1 & 1081.8\\
 & 2 & 59 &  & 5 & 0 &  & 4 & 31.8 &  & 4 & 4 & 1.95 & 1.95 & 1 & 243.8\\
 & 3 & 55 &  & 3 & 0 &  & 3 & 33.8 &  & 3 & 2 & 0.89 & 2 & 381 & *\\
 & 4 & 51 &  & 3 & 0 &  & 3 & 33.7 &  & 3 & 3 & 1.43 & 1.43 & 1 & 223.1\\
 & 5 & 61 &  & 4 & 0 &  & 3 & 53.9 &  & 3 & 3 & 1.72 & 1.72 & 1 & 127.6\\
 & 6 & 55 &  & 4 & 0 &  & 3 & 44.4 &  & 3 & 3 & 1.61 & 1.61 & 1 & 355\\
 & 7 & 63 &  & 5 & 0 &  & 4 & 41.4 &  & 4 & 3 & 1.62 & 3 & 28 & *\\
 & 8 & 43 &  & 3 & 0 &  & 2 & 26.5 &  & 2 & 2 & 1.39 & 1.39 & 1 & 11.1\\
 & 9 & 61 &  & 4 & 0 &  & 3 & 48.8 &  & 3 & 3 & 1.4 & 1.4 & 1 & 894.3\\
 \cmidrule{2-16}
 & average & 57.4 &  & 4.1 & 0 &  & 3.3 & 40.9 &  & 3.3 & 3 & 1.64 & 2.04 & 44.6 & 1373.7\\
\midrule 
25 & 0 & 68 &  & 2 & 0 &  & 2 & 61.3 &  & 2 & 2 & 0.9 & 0.9 & 1 & 341\\
 & 1 & 84 &  & 6 & 0 &  & 5 & 71.9 &  & 5 & 3.86 & 2.46 & 3.86 & 1 & *\\
 & 2 & 50 &  & 3 & 0 &  & 3 & 41.2 &  & 3 & 2 & 1.19 & 2 & 221 & *\\
 & 3 & 80 &  & 6 & 0 &  & 4 & 66.9 &  & 4 & 2.08 & 2.08 & 2.08 & 1 & *\\
 & 4 & 66 &  & 5 & 0 &  & 4 & 44 &  & 4 & 2.47 & 1.33 & 2.47 & 12 & *\\
 & 5 & 72 &  & 4 & 0 &  & 4 & 70.1 &  & 4 & 3 & 1.8 & 3 & 37 & *\\
 & 6 & 74 &  & 4 & 0 &  & 4 & 68 &  & 4 & 4 & 2.37 & 2.37 & 1 & 496.5\\
 & 7 & 62 &  & 3 & 0 &  & 3 & 45 &  & 3 & 2 & 1.17 & 2 & 56 & *\\
 & 8 & 64 &  & 8 & 0 &  & 6 & 63.5 &  & 5 & 5 & 2.13 & 4.03 & 1 & 2668.5\\
 & 9 & 66 &  & 5 & 0 &  & 5 & 65.7 &  & 5 & 4 & 2.13 & 4 & 1 & *\\
 \cmidrule{2-16}
 & average & 68.6 &  & 4.6 & 0 &  & 4 & 59.8 &  & 3.9 & 3.04 & 1.76 & 2.67 & 33.2 & 2870.6\\
\midrule
30 & 0 & 79 &  & 4 & 0 &  & 4 & 96.5 &  & 4 & 3 & 1.51 & 3 & 1 & *\\
 & 1 & 83 &  & 4 & 0 &  & 4 & 80.4 &  & 4 & 2.8 & 1.64 & 2.8 & 1 & *\\
 & 2 & 79 &  & 4 & 0 &  & 4 & 73.9 &  & 4 & 3 & 1.51 & 3 & 1 & *\\
 & 3 & 101 &  & 5 & 0 &  & 5 & 125.9 &  & 4 & 4 & 1.88 & 2.59 & 1 & *\\
 & 4 & 75 &  & 5 & 0 &  & 5 & 82.8 &  & 5 & 3.95 & 2.5 & 3.95 & 1 & *\\
 & 5 & 75 &  & 5 & 0 &  & 4 & 87.7 &  & 4 & 4 & 2.44 & 2.44 & 1 & 557.6\\
 & 6 & 101 &  & 9 & 0 &  & 7 & 131.4 &  & 7 & 2.38 & 2.38 & 2.38 & 1 & *\\
 & 7 & 99 &  & 6 & 0 &  & 5 & 122.8 &  & 5 & 2.01 & 2.01 & 2.01 & 1 & *\\
 & 8 & 79 &  & 6 & 0 &  & 5 & 93.6 &  & 5 & 3.18 & 2.02 & 3.18 & 8 & *\\
 & 9 & 83 &  & 4 & 0 &  & 4 & 91.1 &  & 3 & 3 & 1.37 & 1.37 & 1 & 847.2\\
 \cmidrule{2-16}
 & average & 85.4 &  & 5.2 & 0 &  & 4.7 & 98.6 &  & 4.5 & 3.13 & 1.93 & 2.67 & 1.7 & 3020.4\\
\midrule
35 & 0 & 94 &  & 5 & 0 &  & 4 & 132.6 &  & 4 & 4 & 1.99 & 1.99 & 1 & 2809.1\\
 & 1 & 92 &  & 3 & 0 &  & 3 & 104 &  & 3 & 3 & 1.57 & 1.57 & 1 & 610.6\\
 & 2 & 106 &  & 3 & 0 &  & 3 & 138.6 &  & 3 & 3 & 1.36 & 1.36 & 1 & 1178.4\\
 & 3 & 112 &  & 8 & 0 &  & 7 & 151.6 &  & 7 & 5.41 & 3.82 & 5.41 & 1 & *\\
 & 4 & 86 &  & 5 & 0 &  & 4 & 85 &  & 4 & 4 & 2.44 & 2.44 & 1 & 819.9\\
 & 5 & 104 &  & 5 & 0 &  & 5 & 154.8 &  & 5 & 2.86 & 2.03 & 2.86 & 1 & *\\
 & 6 & 92 &  & 4 & 0 &  & 4 & 104.3 &  & 4 & 1.54 & 1.54 & 1.54 & 1 & *\\
 & 7 & 84 &  & 6 & 0 &  & 6 & 95.4 &  & 6 & 4.59 & 2.82 & 4.59 & 1 & *\\
 & 8 & 94 &  & 5 & 0 &  & 4 & 149.1 &  & 4 & 2.63 & 2.63 & 2.63 & 1 & *\\
 & 9 & 80 &  & 4 & 0 &  & 4 & 92.5 &  & 4 & 3 & 1.84 & 3 & 1 & *\\
 \cmidrule{2-16}
 & average & 94.4 &  & 4.8 & 0 &  & 4.4 & 120.8 &  & 4.4 & 3.4 & 2.2 & 2.74 & 1 & 2701.8\\
\bottomrule
\end{tabular}
\caption{Results for type A instances with $n \leq 35$.}
\label{tab:A}
\end{table}

\begin{table}[htbp]
\centering
\scriptsize
\begin{tabular}{@{}lrrrrrrrrrrrrrrr@{}} 
\toprule
 \multicolumn{3}{c}{Instance} & & \multicolumn{2}{c}{Constructive} & & \multicolumn{2}{c}{ACO} & & \multicolumn{6}{c}{Model}\\
\cmidrule{1-3} \cmidrule{5-6} \cmidrule{8-9} \cmidrule{11-16}
    $n$ & id & tasks & & ub & t & & ub & t & & ub & lb & LR & root & nodes & t\\
\midrule 
10 & 0 & 23 &  & 2 & 0 &  & 2 & 8.8 &  & 2 & 2 & 0.67 & 0.67 & 1 & 3.1\\
 & 1 & 35 &  & 3 & 0 &  & 3 & 12 &  & 3 & 3 & 1.29 & 1.29 & 1 & 43.4\\
 & 2 & 31 &  & 2 & 0 &  & 2 & 13.6 &  & 2 & 2 & 0.75 & 0.75 & 1 & 14.5\\
 & 3 & 35 &  & 4 & 0 &  & 3 & 12.9 &  & 3 & 3 & 1.4 & 1.4 & 1 & 80.8\\
 & 4 & 23 &  & 2 & 0 &  & 1 & 8.5 &  & 1 & 1 & 0.67 & 0.67 & 1 & 0.5\\
 & 5 & 23 &  & 3 & 0 &  & 3 & 8.4 &  & 2 & 2 & 0.87 & 0.87 & 1 & 4.1\\
 & 6 & 29 &  & 3 & 0 &  & 2 & 11.8 &  & 2 & 2 & 0.79 & 0.79 & 1 & 6.9\\
 & 7 & 29 &  & 7 & 0 &  & 6 & 10.9 &  & 6 & 2.86 & 1.45 & 2.83 & 1452 & *\\
 & 8 & 31 &  & 2 & 0 &  & 2 & 15.4 &  & 2 & 2 & 0.83 & 0.83 & 1 & 4.2\\
 & 9 & 19 &  & 1 & 0 &  & 1 & 5.6 &  & 1 & 1 & 0.45 & 0.45 & 1 & 0.2\\
 \cmidrule{2-16}
 & average & 27.8 &  & 2.9 & 0 &  & 2.5 & 10.8 &  & 2.4 & 2.08 & 0.92 & 1.05 & 146.1 & 375.8\\
\midrule
15 & 0 & 48 &  & 2 & 0 &  & 2 & 34 &  & 2 & 2 & 0.99 & 0.99 & 1 & 41.5\\
 & 1 & 48 &  & 4 & 0 &  & 3 & 22.7 &  & 3 & 3 & 1.51 & 1.51 & 1 & 128.8\\
 & 2 & 48 &  & 2 & 0 &  & 2 & 18.2 &  & 2 & 2 & 0.7 & 0.7 & 1 & 40.9\\
 & 3 & 52 &  & 3 & 0 &  & 3 & 28.6 &  & 3 & 2 & 0.97 & 2 & 343 & *\\
 & 4 & 40 &  & 2 & 0 &  & 2 & 17.6 &  & 2 & 2 & 0.9 & 0.9 & 1 & 19.6\\
 & 5 & 42 &  & 3 & 0 &  & 2 & 23.7 &  & 2 & 2 & 0.94 & 0.94 & 1 & 40.1\\
 & 6 & 38 &  & 4 & 0 &  & 3 & 18.1 &  & 3 & 3 & 1.63 & 1.63 & 1 & 56.1\\
 & 7 & 46 &  & 7 & 0 &  & 6 & 20.3 &  & 6 & 2.88 & 1.47 & 2.73 & 38 & *\\
 & 8 & 46 &  & 3 & 0 &  & 2 & 33.2 &  & 2 & 2 & 1.4 & 1.4 & 1 & 14.8\\
 & 9 & 38 &  & 2 & 0 &  & 2 & 18.2 &  & 2 & 1 & 0.61 & 1 & 5885 & *\\
 \cmidrule{2-16}
 & average & 44.6 &  & 3.2 & 0 &  & 2.7 & 23.5 &  & 2.7 & 2.18 & 1.11 & 1.38 & 627.3 & 1114.2\\
\midrule
20 & 0 & 63 &  & 5 & 0 &  & 4 & 48.1 &  & 4 & 4 & 1.9 & 1.9 & 1 & 236.4\\
 & 1 & 63 &  & 7 & 0 &  & 6 & 31.2 &  & 6 & 4.69 & 2.79 & 4.69 & 42 & *\\
 & 2 & 59 &  & 3 & 0 &  & 3 & 30.1 &  & 3 & 1.75 & 0.94 & 1.75 & 165 & *\\
 & 3 & 55 &  & 3 & 0 &  & 2 & 33.6 &  & 2 & 2 & 1.4 & 1.4 & 1 & 29.5\\
 & 4 & 51 &  & 4 & 0 &  & 4 & 33.9 &  & 4 & 3 & 1.51 & 3 & 332 & *\\
 & 5 & 61 &  & 4 & 0 &  & 4 & 43.3 &  & 4 & 3 & 1.65 & 3 & 99 & *\\
 & 6 & 55 &  & 5 & 0 &  & 4 & 43.3 &  & 4 & 3 & 1.52 & 3 & 209 & *\\
 & 7 & 63 &  & 4 & 0 &  & 3 & 42.1 &  & 3 & 3 & 1.82 & 1.82 & 1 & 739.9\\
 & 8 & 43 &  & 3 & 0 &  & 3 & 27 &  & 3 & 3 & 1.35 & 1.35 & 1 & 114.2\\
 & 9 & 61 &  & 10 & 0 &  & 7 & 43.3 &  & 7 & 4.63 & 2.4 & 4.63 & 1 & *\\
 \cmidrule{2-16}
 & average & 57.4 &  & 4.8 & 0 &  & 4 & 37.6 &  & 4 & 3.2 & 1.73 & 2.65 & 85.2 & 2272\\
\midrule
25 & 0 & 68 &  & 9 & 0 &  & 7 & 53.5 &  & 7 & 4.21 & 2.38 & 4.21 & 1 & *\\
 & 1 & 84 &  & 10 & 0 &  & 7 & 63.9 &  & 7 & 3.5 & 3.5 & 3.5 & 1 & *\\
 & 2 & 50 &  & 3 & 0 &  & 3 & 28 &  & 3 & 2 & 1.09 & 2 & 422 & *\\
 & 3 & 80 &  & 5 & 0 &  & 4 & 65.4 &  & 4 & 4 & 1.89 & 1.89 & 1 & 2515.1\\
 & 4 & 66 &  & 4 & 0 &  & 3 & 48.7 &  & 3 & 3 & 1.35 & 1.35 & 1 & 165.3\\
 & 5 & 72 &  & 5 & 0 &  & 4 & 60.8 &  & 4 & 4 & 2.06 & 2.06 & 1 & 3356.6\\
 & 6 & 74 &  & 9 & 0 &  & 7 & 60.2 &  & 6 & 3.3 & 2.11 & 3.3 & 1 & *\\
 & 7 & 62 &  & 3 & 0 &  & 2 & 39.3 &  & 2 & 2 & 1.04 & 1.04 & 1 & 43.5\\
 & 8 & 64 &  & 4 & 0 &  & 3 & 61.5 &  & 3 & 3 & 1.88 & 1.88 & 1 & 410.3\\
 & 9 & 66 &  & 4 & 0 &  & 4 & 64.6 &  & 4 & 3 & 1.79 & 3 & 50 & *\\
 \cmidrule{2-16}
 & average & 68.6 &  & 5.6 & 0 &  & 4.4 & 54.6 &  & 4.3 & 3.2 & 1.91 & 2.42 & 48 & 2449.1\\
\midrule
30 & 0 & 79 &  & 8 & 0 &  & 6 & 83.6 &  & 6 & 4.32 & 2.44 & 4.32 & 1 & *\\
 & 1 & 83 &  & 4 & 0 &  & 4 & 73.9 &  & 4 & 2.68 & 1.54 & 2.68 & 1 & *\\
 & 2 & 79 &  & 4 & 0 &  & 4 & 51.2 &  & 4 & 2.22 & 1.33 & 2.22 & 4 & *\\
 & 3 & 101 &  & 15 & 0 &  & 13 & 117.3 &  & 12 & 2.58 & 2.58 & 2.58 & 1 & *\\
 & 4 & 75 &  & 6 & 0 &  & 5 & 78.5 &  & 5 & 3.76 & 2.22 & 3.76 & 3 & *\\
 & 5 & 75 &  & 8 & 0 &  & 6 & 74.7 &  & 6 & 3.8 & 2.8 & 3.8 & 1 & *\\
 & 6 & 101 &  & 28 & 0 &  & 27 & 99.8 &  & - & - & - & - & - & *\\
 & 7 & 99 &  & 5 & 0 &  & 5 & 122.1 &  & 5 & 1.76 & 1.76 & 1.76 & 1 & *\\
 & 8 & 79 &  & 5 & 0 &  & 4 & 83.3 &  & 4 & 2.04 & 2.04 & 2.04 & 1 & *\\
 & 9 & 83 &  & 4 & 0 &  & 4 & 81.6 &  & 4 & 2 & 1.46 & 2 & 1 & *\\
 \cmidrule{2-16}
 & average & 83.6 &  & 6.5 & 0 &  & 5.6 & 85.1 &  & 5.5 & 2.79 & 2.02 & 2.79 & 1.5 & 3600\\
\midrule
35 & 0 & 94 &  & 10 & 0 &  & 8 & 124.6 &  & 7 & 2.78 & 2.78 & 2.78 & 1 & *\\
 & 1 & 92 &  & 6 & 0 &  & 4 & 59.5 &  & 4 & 2.34 & 2.34 & 2.34 & 1 & *\\
 & 2 & 106 &  & 5 & 0 &  & 4 & 102 &  & 4 & 1.88 & 1.88 & 1.88 & 1 & *\\
 & 3 & 112 &  & 10 & 0 &  & 7 & 107.4 &  & 7 & 3.52 & 3.52 & 3.52 & 1 & *\\
 & 4 & 86 &  & 4 & 0 &  & 3 & 86.7 &  & 3 & 3 & 1.55 & 1.55 & 1 & 2019.6\\
 & 5 & 104 &  & 9 & 0 &  & 7 & 114.6 &  & 7 & 3.58 & 2.89 & 3.58 & 1 & *\\
 & 6 & 92 &  & 4 & 0 &  & 4 & 113.8 &  & 4 & 1.77 & 1.77 & 1.77 & 1 & *\\
 & 7 & 84 &  & 6 & 0 &  & 5 & 70.2 &  & 5 & 2.36 & 2.36 & 2.36 & 1 & *\\
 & 8 & 94 &  & 21 & 0 &  & 16 & 101.5 &  & - & - & - & - & - & *\\
 & 9 & 80 &  & 8 & 0 &  & 7 & 78.1 &  & 7 & 3.08 & 2.58 & 3.08 & 1 & *\\
 \cmidrule{2-16}
 & average & 94.4 &  & 6.8 & 0 &  & 5.4 & 95.2 &  & 5.3 & 2.7 & 2.41 & 2.54 & 1 & 3424.4\\
\bottomrule
\end{tabular}
\caption{Results for type B instances with $n \leq 35$.}
\label{tab:B}
\end{table}

\begin{table}[htbp]
\centering
\scriptsize
\begin{tabular}{@{}lrrrrrrrrrrrrrrr@{}} 
\toprule
 \multicolumn{3}{c}{Instance} & & \multicolumn{2}{c}{Constructive} & & \multicolumn{2}{c}{ACO} & & \multicolumn{6}{c}{Model}\\
\cmidrule{1-3} \cmidrule{5-6} \cmidrule{8-9} \cmidrule{11-16}
    $n$ & id & tasks & & ub & t & & ub & t & & ub & lb & LR & root & nodes & t\\
\midrule 
10 & 0 & 27 &  & 4 & 0 &  & 4 & 10.2 &  & 3 & 3 & 1.26 & 2.36 & 1 & 11.6\\
 & 1 & &  & 10 & 0 &  & 10 & 11.3 &  & 10 & 5.02 & 1.91 & 5.02 & 766 & *\\
 & 2 & &  & 9 & 0 &  & 9 & 12.8 &  & 9 & 3.98 & 2 & 3.93 & 909 & *\\
 & 3 & &  & 10 & 0 &  & 10 & 8.5 &  & 10 & 8.99 & 1.93 & 8.7 & 716 & *\\
 & 4 & &  & 10 & 0 &  & 10 & 10 &  & 9 & 4.23 & 1.93 & 4.09 & 715 & *\\
 & 5 & &  & 2 & 0 &  & 2 & 7.5 &  & 2 & 2 & 1.6 & 1.6 & 1 & 1.9\\
 & 6 & &  & 3 & 0 &  & 3 & 7 &  & 3 & 2 & 0.97 & 2 & 24101 & *\\
 & 7 & &  & 5 & 0 &  & 5 & 10.1 &  & 5 & 4 & 1.98 & 4 & 4438 & *\\
 & 8 & &  & 5 & 0 &  & 5 & 8.8 &  & 5 & 3.19 & 1.96 & 3.14 & 3048 & *\\
 & 9 & &  & 2 & 0 &  & 2 & 9.9 &  & 2 & 2 & 0.61 & 1 & 1889 & 277\\
 \cmidrule{2-16}
 &  & 27 &  & 6 & 0 &  & 6 & 9.6 &  & 5.8 & 3.84 & 1.62 & 3.58 & 3658.4 & 2549\\
\midrule
15 & 0 & 42 &  & 4 & 0 &  & 5 & 21.1 &  & 3 & 3 & 1.39 & 2.8 & 1 & 192.3\\
 & 1 &  &  & 15 & 0 &  & 15 & 24.1 &  & 14 & 6.79 & 2.92 & 6.74 & 61 & *\\
 & 2 &  &  & 14 & 0 &  & 14 & 26.4 &  & 14 & 5.63 & 2.99 & 5.61 & 55 & *\\
 & 3 &  &  & 5 & 0 &  & 5 & 15 &  & 5 & 3.79 & 2.13 & 3.78 & 494 & *\\
 & 4 &  &  & 4 & 0 &  & 3 & 19.4 &  & 3 & 2 & 1.02 & 2 & 2216 & *\\
 & 5 & &  & 14 & 0 &  & 14 & 15.6 &  & 14 & 5.85 & 3.01 & 5.85 & 63 & *\\
 & 6 &  &  & 15 & 0 &  & 15 & 19.1 &  & 15 & 5.85 & 2.96 & 5.85 & 43 & *\\
 & 7 &  &  & 15 & 0 &  & 15 & 20.6 &  & 14 & 5.62 & 2.88 & 5.62 & 51 & *\\
 & 8 &  &  & 7 & 0 &  & 7 & 19.6 &  & 7 & 5.49 & 3.27 & 5.49 & 263 & *\\
 & 9 &  &  & 6 & 0 &  & 5 & 18.3 &  & 5 & 3.76 & 2.2 & 3.76 & 481 & *\\
 \cmidrule{2-16}
 &  & 42 &  & 9.9 & 0 &  & 9.8 & 19.9 &  & 9.4 & 4.78 & 2.48 & 4.75 & 372.8 & 3259.2\\
\bottomrule
\end{tabular}
\caption{Results for type C instances with $n \in \{10, 15\}$.}
\label{tab:C}
\end{table}

In the second set of experiments, we evaluate the performance of the heuristic algorithms on larger instances. For these experiments, we also consider a version of the MMAS ACO algorithm using 10 instead of 100 ants. The results are summarized in Table \ref{tab:heu}. Under ``Instance'', the table presents instance information. For each algorithm, the table presents the average upper bound (ub) and computational time (t) in seconds over the 10 instances for the type and instance size in that row. For the ACO implementations, the table also presents the number of wins and losses in relation to the constructive heuristic.

\begin{figure} 
\begin{center} 
\includegraphics[width=14cm]{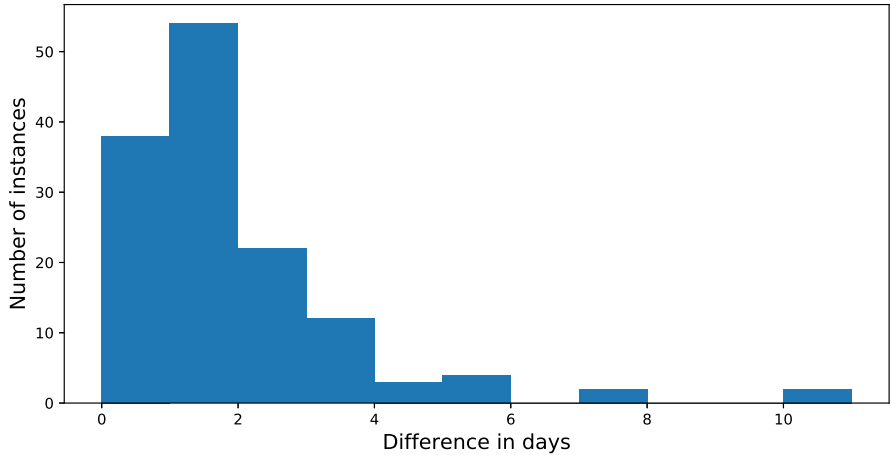}
\caption{Difference in days between the solution found by the constructive heuristic and the solution found by the MMAS ACO algorithm with 100 ants, for type A and B instances.}  
\label{fig:hist}
 \end{center} 
 \end{figure}

In type A and type B instances, the MMAS ACO algorithms significantly outperform the constructive heuristic. The implementation with 100 ants finds strictly better solutions in 47 and 52 out 70 type A and 70 type B instances, respectively. The implementation with 10 ants is significantly faster, and still manages to find strictly better solutions than the constructive heuristic in 38 and 49, out of 70 type A and 70 type B instances, respectively. However, there is a decrease in solution quality in relation to the implementation with 100 ants, the implementation with 10 ants failed to match the best solution found by that implementation in 40 out of 140 type A and B instances. In Figure \ref{fig:hist}, we present a histogram with the difference in days between the solutions found by the ACO implementation with 100 ants and the constructive heuristic, in type A and B instances. In 45 cases, the difference was 2 days or more. In 23 cases, the difference was 3 days or more. In 2 cases, the difference was 10 days. 

In type C instances, the performance of the ACO algorithms is more modest. Compared to the constructive heuristic, the ACO implementation with 100 ants found a better solution in in 7, a worse solution in 4, and the same solution in 99 out of 110 instances. The implementation with 10 ants was better in 5 cases, worse in 8, and found the same solution in 97 instances.

Overall, the MMAS ACO implementation seems to be an appropriate approach for solving the MWSRPDT.

\begin{table}[htbp]
\centering
\scriptsize
\begin{tabular}{@{}lrrrrrrrrrrrrrrr@{}} 
\toprule
    & & & & & & & \multicolumn{9}{c}{ACO}\\ 
\cmidrule{8-16}
\multicolumn{3}{c}{Instance} & & \multicolumn{2}{c}{Constructive} & & \multicolumn{4}{c}{10 Ants} & & \multicolumn{4}{c}{100 Ants}\\ 
\cmidrule{1-3} \cmidrule{5-6} \cmidrule{8-11} \cmidrule{13-16}
type & $n$ & tasks & & ub & t & & wins & losses & ub & t & & wins & losses & ub & t\\
\midrule 
A & 40 & 121.2 &  & 8 & 0 &  & 8 & 0 & 7 & 34.1 &  & 9 & 0 & 6.7 & 176.8\\
 & 50 & 145.2 &  & 7.5 & 0 &  & 3 & 0 & 7.1 & 59.8 &  & 4 & 0 & 7 & 275.9\\
 & 60 & 171.2 &  & 10.9 & 0 &  & 6 & 0 & 10.1 & 85.7 &  & 8 & 0 & 9.8 & 390.5\\
 & 70 & 201.2 &  & 9.4 & 0 &  & 5 & 1 & 8.8 & 134.5 &  & 6 & 0 & 8.5 & 566.3\\
 & 80 & 235 &  & 11.8 & 0 &  & 4 & 1 & 11.4 & 197.4 &  & 5 & 1 & 11.2 & 801\\
 & 90 & 270.6 &  & 16.1 & 0 &  & 6 & 0 & 15 & 239.1 &  & 7 & 0 & 14.7 & 1003.9\\
 & 100 & 304 &  & 18.3 & 0 &  & 6 & 0 & 17.5 & 344.8 &  & 8 & 0 & 17.2 & 1289.4\\
\midrule
B & 40 & 121.2 &  & 8.3 & 0 &  & 8 & 0 & 7.3 & 31.6 &  & 8 & 0 & 7.1 & 165.5\\
 & 50 & 145.2 &  & 9.2 & 0 &  & 9 & 0 & 7.4 & 45.1 &  & 9 & 0 & 7.1 & 225.7\\
 & 60 & 171.2 &  & 16.1 & 0 &  & 7 & 1 & 14.3 & 62.9 &  & 9 & 0 & 13.7 & 328.5\\
 & 70 & 201.2 &  & 14.8 & 0 &  & 7 & 0 & 13 & 94 &  & 7 & 0 & 12.7 & 435.6\\
 & 80 & 235 &  & 18.4 & 0 &  & 8 & 0 & 15.9 & 155.3 &  & 9 & 0 & 15.5 & 672.2\\
 & 90 & 270.6 &  & 15.8 & 0 &  & 6 & 1 & 14.5 & 189.1 &  & 6 & 0 & 14.1 & 805.1\\
 & 100 & 304 &  & 15.5 & 0 &  & 4 & 2 & 15.4 & 290.6 &  & 4 & 2 & 15.2 & 1179.5\\
\midrule
C & 20 & 57 &  & 9.1 & 0 &  & 1 & 0 & 9 & 3.5 &  & 1 & 0 & 9 & 31\\
 & 25 & 72 &  & 11.3 & 0 &  & 0 & 0 & 11.3 & 5.4 &  & 0 & 0 & 11.3 & 45.6\\
 & 30 & 87 &  & 18.6 & 0 &  & 1 & 0 & 18.5 & 8.6 &  & 1 & 0 & 18.5 & 71.2\\
 & 35 & 102 &  & 13.6 & 0 &  & 0 & 0 & 13.6 & 12.1 &  & 1 & 0 & 13.5 & 86.5\\
 & 40 & 117 &  & 16.3 & 0 &  & 1 & 0 & 16.2 & 14.4 &  & 1 & 0 & 16.2 & 108.4\\
 & 50 & 147 &  & 18.2 & 0 &  & 1 & 3 & 18.4 & 24.9 &  & 2 & 1 & 18.1 & 166.3\\
 & 60 & 177 &  & 39.6 & 0 &  & 0 & 0 & 39.6 & 40.7 &  & 0 & 0 & 39.6 & 272.2\\
 & 70 & 207 &  & 27.9 & 0 &  & 0 & 1 & 28 & 62 &  & 0 & 1 & 28 & 333.3\\
 & 80 & 237 &  & 41.5 & 0 &  & 0 & 1 & 41.6 & 84 &  & 0 & 0 & 41.5 & 486.7\\
 & 90 & 267 &  & 51.1 & 0 &  & 1 & 1 & 51.1 & 114.6 &  & 1 & 1 & 51.1 & 605.7\\
 & 100 & 297 &  & 53.3 & 0 &  & 0 & 2 & 53.9 & 134.2 &  & 0 & 1 & 53.7 & 737.8\\
\bottomrule
\end{tabular}
\caption{Results for the heuristic algorithms on larger instances.}
\label{tab:heu}
\end{table}

\section{Conclusion}
\label{sec:conclusions}

In this work, we introduced a new Workforce Scheduling and Routing Problem denoted Multiperiod Workforce Scheduling and Routing Problem with Dependent Tasks. In this problem, a company provides services composed of interdependent tasks to customers. The problem consists in assigning tasks to mobile teams throughout the minimum possible number of days.

We introduced a Mixed-Integer Programming model for the problem that is capable of consistently solving problems with up to 20 customers. In order to obtain solutions in a small amount of time and solve larger problems, we introduced a constructive heuristic based on Discrete Event Simulation. Since different teams might visit customers on a single day, one has to be careful with the synchronization of teams so that dependencies are not broken. This also poses a problem for the development of local search algorithms, as even small changes might make the solution infeasible. This characteristic of the problem makes the ACO framework ideal for the development of good algorithms, as it is capable of learning what constitutes a high-quality solution without necessarily employing local search. The constructive heuristic was used as the basis for the development of algorithms based on the three main ACO variations. Each variation was tested with four types of problem components. The winning ACO algorithm, an MMAS ACO algorithm using components of the type $(h, k,(i,a), (j,b))$, where $h$ is a day, $k$ is a team, and $(i,a)$ and $(j,b)$ are customer-task pairs, was demonstrated to generate high quality solutions in a small amount of time.

In the future, we intend to investigate stronger models, such as the ones based on set covering, as well as the application of other metaheuristic frameworks.

\section*{Acknowledgements}
This work was financed by FAPEMIG grant APQ-02762-17.

\bibliographystyle{plainnat}
\bibliography{references}

\end{document}